\documentclass{article}

     \usepackage[preprint, nonatbib]{neurips_2019}

\usepackage[utf8]{inputenc} 
\usepackage[T1]{fontenc}    
\usepackage{hyperref}       
\usepackage{url}            
\usepackage{booktabs}       
\usepackage{amsfonts}       
\usepackage{nicefrac}       
\usepackage{microtype}      

\usepackage{amsmath}
\usepackage{amsthm}
\usepackage{dsfont}
\usepackage{color}
\usepackage{graphicx}

\theoremstyle{definition}

\usepackage[title]{appendix}

\usepackage[numbers]{natbib}

\title{Tree-Transformer: A Transformer-Based Method for Correction of Tree-Structured Data.}

\author{%
  Jacob Harer \\
  Boston University\\
  \texttt{jharer@bu.edu} \\
   \And
  Chris Reale \\
  Draper \\
  \texttt{creale@draper.com} \\
  \AND
  Peter Chin \\
  Boston University \\
  \texttt{spchin@cs.bu.edu} \\
}

\graphicspath{{Figures/}}
\DeclareGraphicsExtensions{.pdf,.jpeg,.png,.jpg} 

\begin{document}

\maketitle

\begin{abstract}
    Many common sequential data sources, such as source code and natural language, have a natural tree-structured representation. These trees can be generated by fitting a sequence to a grammar, yielding a hierarchical ordering of the tokens in the sequence. This structure encodes a high degree of syntactic information, making it ideal for problems such as grammar correction. However, little work has been done to develop neural networks that can operate on and exploit tree-structured data. In this paper we present the Tree-Transformer \textemdash{} a novel neural network architecture designed to translate between arbitrary input and output trees. We applied this architecture to correction tasks in both the source code and natural language domains. On source code, our model achieved an improvement of $25\%$ $\text{F}0.5$ over the best sequential method. On natural language, we achieved comparable results to the most complex state of the art systems, obtaining a $10\%$ improvement in recall on the CoNLL 2014 benchmark and the highest to date $\text{F}0.5$ score on the AESW benchmark of $50.43$.  
\end{abstract}

\section{Introduction}
Most machine learning approaches to correction tasks operate on sequential representations of input and output data. Generally this is done as a matter of convenience \textemdash{} sequential data is readily available and requires minimal effort to be used as training data for many machine learning models. Sequence-based machine learning models have produced prominent results in both translation and correction of natural language \cite{Bahdanau:2014vz, Vaswani:2017ul, Xie:2016vi, Yuan:2016kf, Ji:2017gk, Schmaltz:2017ur, Chollampatt:2018tj, JunczysDowmunt:2018}.

While algorithms that use sequential data have produced good results, most of these sequential data types can be more informatively represented using a tree structure. One common method to obtain tree-structured data is to fit sequential data to a grammar, such as a Context Free Grammar (CFG). The use of a grammar ensures that generated trees encode the higher-order syntactic structure of the data in addition to the information contained by sequential data.

In this work, we trained a neural network to operate directly on trees, teaching it to learn the syntax of the underlying grammar and to leverage this syntax to produce outputs which are grammatically correct. Our model, the Tree-Transformer, handles correction tasks in a tree-based encoder and decoder framework. The Tree-Transformer leverages the popular Transformer architecture \cite{Vaswani:2017ul}, modifying it to incorporate the tree structure of the data by adding a parent-sibling tree convolution block. To show the power of our model, we focused our experiments on two common data types and their respective tree representations: Abstract Syntax Trees (ASTs) for code and Constituency Parse Trees (CPTs) for natural language. 




\section{Related Work}

\subsection{Tree Structured Neural Networks}
Existing work on tree-structured neural networks can largely be grouped into two categories: encoding trees and generating trees. Several types of tree encoders exist \cite{Tai:2015wp, Zhu:2015vo, Socher:2011wx, Eriguchi:2016ub}. The seminal work of Tai et al. \cite{Tai:2015wp} laid the ground work for these methods, using a variant of a Long Short Term Memory (LSTM) network to encode an arbitrary tree. A large body of work has also focused on how to generate trees \cite{Dong:2016vv, Alvarez:2017, Vinyals:2014uh, Aharoni:2017um, Rabinovich:2017ut, Parisotto:2016te, Yin:2017vb, Zhang:2015uv}. The work of Dong et al., \cite{Dong:2016vv} and Alvarez-Melis and Jaakkola \cite{Alvarez:2017} each extend the LSTM decoder popular in Neuro Machine Translation (NMT) systems to arbitrary trees. This is done by labeling some outputs as parent nodes and then forking off additional sequence generations to create their children. Only a small amount of work has combined encoding and generation of trees into a tree-to-tree system \cite{Chen:2018wd, Chakraborty:2018}. Of note is Chakraborty et al. \cite{Chakraborty:2018} who use a LSTM-based tree-to-tree method for source code completion.  


To our knowledge our work is the first to use a Transformer-based network on trees, and apply Tree-to-Tree techniques to natural language and code correction tasks. 

\subsection{Code Correction}
There is extensive existing work on automatic repair of software. However, the majority of this work is rule-based systems which make use of small datasets (see \cite{Monperrus:2018:survey} for a more extensive review of these methods). Two successful, recent approaches in this category are that of Le et al., \cite{Le:2016history} and Long and Rinard \cite{Long:2016automatic}. Le et al. mine a history of bug fixes across multiple projects and attempt to reuse common bug fix patterns on newly discovered bugs. Long and Rinard learn and use a probabilistic model to rank potential fixes for defective code. Unfortunately, the small datasets used in these works are not suitable for training a large neural network like ours. 

Neural network-based approaches for code correction are less common. Devlin et al. \cite{Devlin:2017tf} generate repairs with a rule-based method and then rank them using a neural network. Gupta et al. \cite{Gupta:2017deepfix} were the first to train a NMT model to directly generate repairs for incorrect code. Additionally, Harer et al. \cite{Harer:2018wo} use a Generative Adversarial Network to train an NMT model for code correction in the absence of paired data. The works of Gupta et al. and Harer et al. are the closest to our own since they directly correct code using an NMT system. 

\subsection{Grammatical Error Correction}
Grammatical Error Correction (GEC) is the task of correcting grammatically incorrect sentences. This task is similar in many ways to machine translation tasks. However, initial attempts to apply NMT systems to GEC were outperformed by phrase-based or hybrid systems \cite{JunczysDowmunt:2016um, Chollampatt:2017il, Dahlmeier:2013ui}. 

Initial, purely neural systems for GEC largely copied NMT systems. Yuan and Brisco \cite{Yuan:2016kf} produced the first NMT style system for GEC by using the popular attention method of Bahdanau et al. \cite{Bahdanau:2014vz}. Xie et al. \cite{Xie:2016vi} trained a novel character-based model with attention. Ji et al. \cite{Ji:2017gk} proposed a hybrid character-word level model, using a nested character level attention model to handle rare words. Schmaltz et al. \cite{Schmaltz:2017ur} used a word-level bidirectional LSTM network. Chollampatt and Ng \cite{Chollampatt:2018tj} created a convolution-based encoder and decoder network which was the first to beat state of the art phrased-based systems. Finally, Junczys-Dowmunt et al. \cite{JunczysDowmunt:2018} treated GEC as a low resource machine translation task, utilizing the combination of a large monolingual language model and a specifically designed correction loss function.

\section{Architecture}
Our Tree-Transformer architecture is based on the Transformer architecture of Vaswani et al. \cite{Vaswani:2017ul}, modified to handle tree-structured data. Our major change to the Transformer is the replacement of the feed forward sublayer in both the encoder and decoder with a Tree Convolution Block (TCB). The TCB allows each node direct access to its parent and left sibling, thus allowing the network to understand the tree structure. We follow the same overall architecture for the Transformer as Vaswani et al., consisting of self-attention, encoder-decoder attention, and TCB sub layers in each layer. Our models follow the 6 layer architecture of the base Transformer model with sublayer outputs, $d_{model}$, of size 512 and tree convolution layers, $d_{ff}$, of size $2048$.

\begin{figure}
\centering
\includegraphics[width=0.7\linewidth]{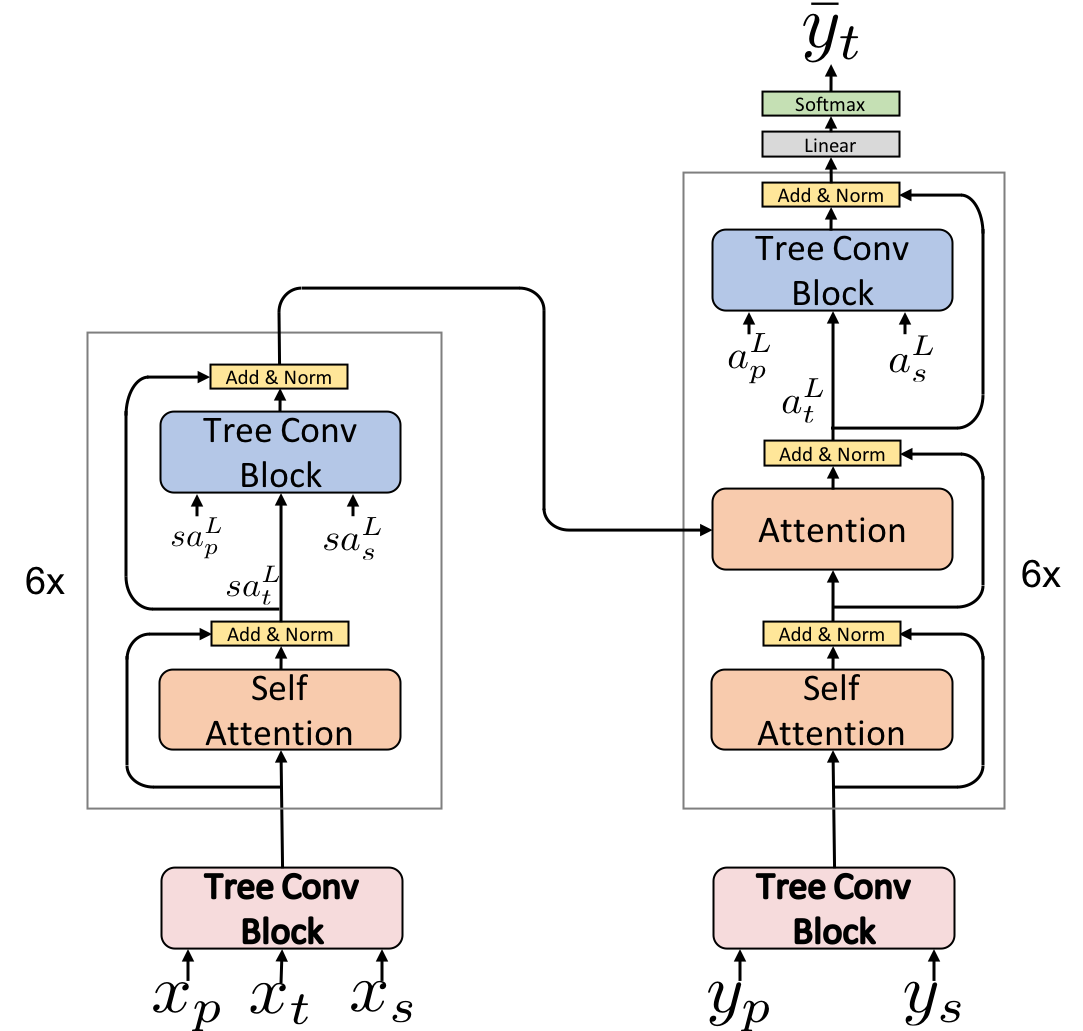}
\caption{Tree-Transformer model architecture.}
\label{fig:model_arch}
\end{figure}

\subsection{Parent-Sibling Tree Convolution}
Tree convolution is computed for each node as:
\[ TCB(x_t, x_p, x_s) = relu(x_tW_t + x_pW_p + x_sW_s + b)W_2 + b_2 \]
The inputs $x_t$, $x_p$, and $x_s$ all come from the previous sublayer, $x_t$ from the same node, $x_p$ from the parent node, and $x_s$ from its left sibling. In cases where the node does not have either a parent (e.g. the root node) or a left sibling (e.g. parents first child), the inputs $x_p$ and $x_s$ are replaced with a learned vector $v_p$ and $v_s$ respectively.

In addition to the TCB used in each sub layer, we also use a TCB at the input to both the encoder and decoder. In the encoder this input block combines the embeddings from the parent, the sibling and the current node ($p$, $s$, and $t$). In the decoder, the current node is unknown since it has not yet been produced. Therefor,e this block only combines the parent and sibling embeddings, leaving out the input $x_t$ from the equation above.

The overall structure of the network is shown in Figure \ref{fig:model_arch}. The inputs to each TCB come from the network inputs, $x/y$, for the input blocks, and from the previous sublayer, $sa^L/a^L$, for all other blocks.

\begin{figure}
\centering
\includegraphics[width=0.7\linewidth]{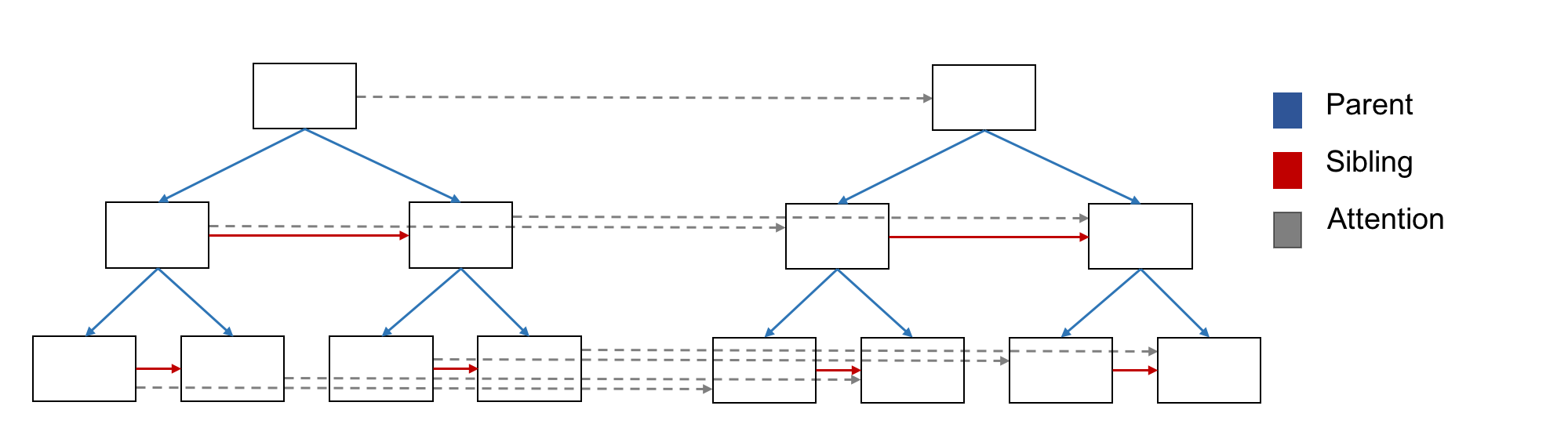}
\caption{Tree-Transformer State Transfer}
\label{fig:state_transfer}
\end{figure}

\subsection{Top-Down Encoder and Decoder}
Both our encoder and decoder use a top-down approach where information flows from the tree's root node down toward to the leaf nodes, as shown in Figure \ref{fig:state_transfer}. Thus, leaf nodes have access to a large sub-tree, while parent nodes have a more limited view. An alternative approach would be to use a bottom-up encoder, where each node has access to its children, and a top-down decoder which can disseminate this information. This bottom-up/top-down model is intuitive because information flows up the tree in the encoder and then back down the decoder. However, we found that utilizing the same top-down ordering for both encoder and decoder performed better, likely because the symmetry in encoder and decoder allows decoder nodes to easily attend to their corresponding nodes in the encoder. This symmetry trivializes copying from encoder to decoder, which is particularly useful in correction tasks where large portions of the trees remain unchanged between input and output.

\subsection{Generating Tree Structure}
In order to generate each tree's structure, we treat each set of siblings as a sequence. For each set of siblings, we generate each node one at a time, ending with the generation of an end-of-sequence token. The vocabulary used defines a set of leaf and parent nodes. When a parent node is generated, we begin creation of that nodes children as another set of siblings.

\subsection{Depth First Ordering} \label{sec:self_attn}

As with any NMT system, each output value $y_t$ is produced from the decoder and fed back to subsequent nodes as input during evaluation. Ensuring inputs $y_p$ and $y_s$ are available to each node requires that parents are produced before children and that siblings are produced in left-to-right order. To enforce this constraint, we order the nodes in a depth-first manner. This ordering is shown by the numbering on nodes in Figure \ref{fig:cond_ex}. The self attention mechanism in the decoder is also masked according to this order, so that each node only has access to previously produced ones.

\subsection{No Positional Encoding}
The Transformer architecture utilizes a positional encoding to help localization of the attention mechanisms. Positional encoding is not as important in our model because the TCB allows nodes to easily locate its parent and siblings in the tree. In fact, we found that inclusion of a Positional Encoding caused the network to overfit, likely due to the relatively small size of the correction datasets we use. Given this, our Tree-Transformer networks do not include a Positional Encoding. 

\section{Why Tree Transformer}

In this section we motivate the design of our Tree-Transformer model over other possible tree-based architectures. Our choice to build upon the Transformer model was two-fold. First, Transformer-based models have significantly reduced time-complexity relative to Recurrent Neural Network (RNN) based approaches. Second, many of the building blocks required for a tree-to-tree translation system, including self-attention, are already present in the Transformer architecture.

\subsection{Recurrent vs Attention Tree Networks} \label{sec:rnn_cnn}
Many previous works on tree-structured networks used RNN-based tree architectures where nodes in layer $L$ are given access to their parent or children in the same layer \cite{Dong:2016vv, Alvarez:2017, Chakraborty:2018}. This state transfer requires an ordering to the nodes during training where earlier nodes in the same layer must be computed prior to later ones. This ordering requirement leads to poor time complexity for tree-structured RNN's, since each node in a tree needs access to multiple prior nodes (e.g. parent and sibling). Accessing the states of prior nodes thus requires a gather operation over all past produced nodes. These gather operations are slow, and performing them serially for each node in the tree can be prohibitively expensive. 

An alternative to the RNN type architecture is a convolutional or attention-based one, where nodes in layer $L$ are given access to prior nodes in layer $L-1$. With the dependence in the same layer removed, the gather operation can be batched over all nodes in a tree, resulting in one large gather operation instead of $T$. From our experiments, this batching resulted in a reduction in training time of two orders of magnitude on our largest dataset; from around $2$ months to less than a day. 

\subsection{Conditional Probabilities and Self Attention} \label{sec:cond_prob}

\begin{figure}
\centering
\includegraphics[width=0.7\linewidth]{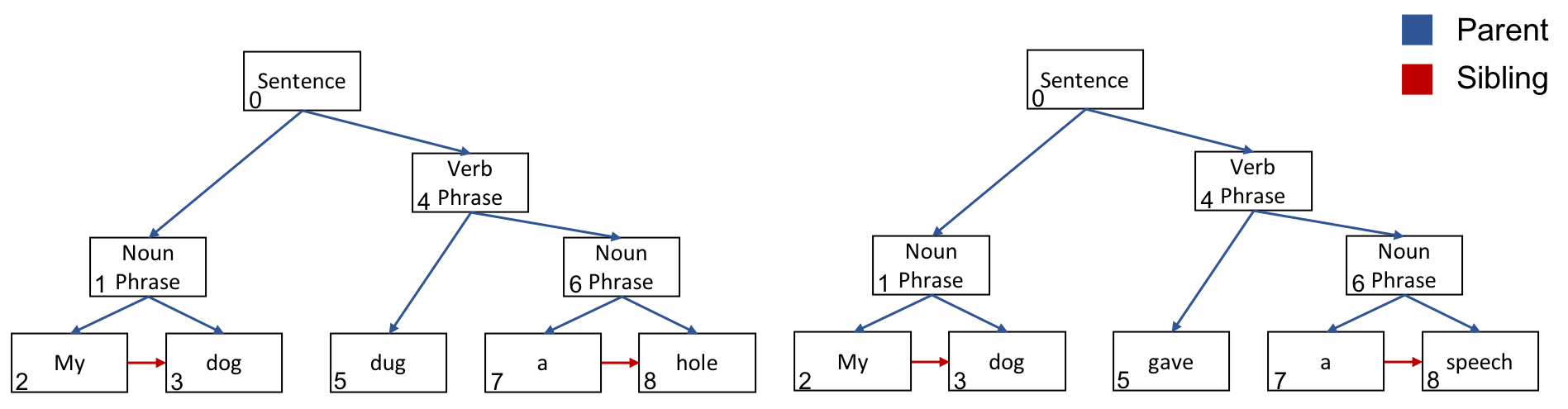}
\caption{Example Constituency Parse Tree. The index of the node in depth-first ordering is shown in the bottom left of each node. Note: leaf nodes in the verb phrase do not have access to leaf nodes in the left noun phrase without self-attention}
\label{fig:cond_ex}
\end{figure}

The Tree-Transformer's structure helps the network produce grammatically correct outputs. However, for translation/correction tasks we must additionally ensure that each output, $\bar{y}_t$, is conditionally dependent on both the input, $x$, and on previous outputs, $y_{<t}$. Conditioning the output on the input is achieved using an encoder-decoder attention mechanism \cite{Bahdanau:2014vz}. Conditioning each output on previous outputs is more difficult with a tree-based system. In a tree-based model like ours, with only parent and sibling connections, the leaf nodes in one branch do not have access to leaf nodes in other branches. This leads to potentially undesired conditional independence between branches. Consider the example constituency parse trees shown in Figure \ref{fig:cond_ex}. Given the initial noun phrase "My dog", the following verb phrase "dug a hole" is far more likely than "gave a speech". However, in a tree-based model the verb phrases do not have direct access to the sampled noun phrase, meaning both possible sentences would be considered roughly equally probable by the model.

We address the above limitation with the inclusion of a self-attention mechanism which allows nodes access to all previously produced nodes. This mechanism, along with the depth-first ordering of the node described in section \ref{sec:self_attn}, gives each leaf node access to all past produced leaf nodes. Our model fits the standard probabilistic language model given as:
\begin{equation} \label{eq:lm}
p(y | x) = \prod_{t=1}^T p(y_t | y_{<t}, x)
\end{equation}
where $t$ is the index of the node in the depth first ordering. 

\section{Training} \label{sec:training}

This section describes the training procedure and parameter choices used in our model. We trained our models in parallel on $4$ Nvidia Tesla V-100 GPU's. Trees were batched together based on size with each batch containing a maximum $20,000$ words. We used the ADAM optimizer with inverse square root decay and a warm up of $16,000$ steps. A full list of hyperparameters for each run is included in Appendix \ref{app:A}. 

\subsection{Regularization}
The correction datasets we used in this paper are relatively small compared to typical NMT datasets. As such we found a high degree of regularization was necessary. We included dropout of 0.3 before the residual of each sub layer and attention dropout of 0.1. We also added dropout of 0.3 to each TCB after the non-linearity. We applied dropout to words in both source and target embeddings as per \cite{JunczysDowmunt:2018} with probabilities 0.2 and 0.1 respectively. We included label smoothing with $\epsilon_{ls} = 0.1$

\subsection{Beam-Search}
Because of the depth-first ordering of nodes in our model, we can use beam search in the same way as traditional NMT systems. Following equation \ref{eq:lm}, we can compute the probability of a generated sub-tree of $t$ nodes simply as the product of probabilities for each node. We utilize beam-search during testing with a beam width of 6. Larger beam widths did not produce improved results.

\section{Experiments/Results}

\subsection{Code Correction}
We trained our Tree-Transformer on code examples taken from the NIST SATE IV dataset \cite{Okun:2013}. SATE IV contains around $120K$ C and C++ files from 116 different Common Weakness Enumerations (CWEs), and was originally designed to test static analyzers. Each file contains a bad function with a known security vulnerability and at least one good function which fixes the vulnerability. We generate Abstract Syntax Trees (ASTs) from these functions using the Clang AST framework \cite{Lattner:2004vw, CLANG}. 

To provide a representation which is usable to our network, we tokenize the AST over a fixed vocabulary in three ways. First, high level AST nodes and data types are represented by individual tokens. Second, character and numeric literals are represented by a sequence of ASCII characters with a parent node defining the kind of literal (e.g. Int Literal, Float Literal). Finally, we use variable renaming to assign per function unique tokens to each variable and string. Our vocabulary consists of $403$ tokens made up of $60$ AST tokens, $23$ data type tokens, $256$ ASCII tokens, and $64$ Variable tokens.

Using the SATE IV dataset requires pre-processing, which we do during data generation. First, many of the SATE IV functions contain large amounts of dead code. In these cases, the bad and good functions contain largely the same code, but one path will be executed during the bad function and another in the good one. To make these cases more realistic, we removed the dead code. Second, although each file for a particular CWE contains unique functions at the text level, many of them are identical once converted to AST with renamed variables. Identical cases comes in two flavors: one where a bad function is identical to its good counterpart, and one where multiple bad functions from different files are identical. The first occurs commonly in SATE IV in cases where the bad and good functions are identical except for differing function calls. Since we operate at the function level, examples of this case are not useful and are removed. The second case occurs when bad functions differ only in variable names, strings, or function calls. To handle this, we compare all bad functions tree representations and combine identical bad functions into a single bad tree and a combination of all good trees from its component functions, with duplicate good trees removed. After pre-processing the data, we retain a total of $32,316$ bad functions and $47,346$ good functions. These are split 80/10/10 into training/validation/testing.

To our knowledge this processing of the SATE IV dataset is new. As such, we compare our network to two NMT systems operating on sequence-based representation of the data; a 4 Layer LSTM with attention and a base Transformer model. These sequence-based models use a representation with an almost identical vocabulary and tokenization to our tree representation but they operate over the tokenized sequence output of the Clang Lexer instead of the AST. 

During testing, we utilized clangs source-to-source compilation framework to return the tree output of our networks to source code. We then compute precision, recall, and $\text{F}0.5$ scores of source code edits using the MaxMatch algorithm \cite{m2scorer:14}. Results are given in Table \ref{tab:sate4} Our Tree-Transformer model performs better than either of the other two models we considered. We believe this is because source code is more naturally structured as a tree than as a sequence, lending itself to our tree-based model. 

\begin{table}
  \vspace{-2.5mm}
  \caption{Sate IV Results}
  \label{tab:sate4}
  \centering
  \begin{tabular}{ l l l l l }
    \toprule
    Architecture & Precision & Recall & F$0.5$ \\
    \midrule
    4-layer LSTM & 51.3 & 53.4 & 51.7\\
    Transformer & 59.6 & 86.1 & 63.5\\
    Tree-Transformer & \textbf{84.5} & \textbf{85.7} & \textbf{84.7} \\
    \bottomrule
  \end{tabular}
  \vspace{-2.5mm}
\end{table}

\subsection{Grammar Error Correction}

We applied our tree-based model as an alternative to NMT and phrase-based methods for GEC. Specifically, we encoded incorrect sentences using their constituency parse trees, and then generated corrected parse trees. Constituency parse trees represent sentences based on their syntactic structure by fitting them to a phrase structured grammar. Words from the input sentence become leaf nodes of their respective parse trees, and these nodes are combined into phrases of more complexity as we progress up the tree. \cite{Goller:jh, Socher:2011wx} 

A large amount of research has focused on the generation of constituency parse trees \cite{Danqi:14, Socher:13, Klein:03}. We utilize the Stanford NLP group's shift-reduce constituency parser \cite{StNLP} to generate trees for both incorrect and correct sentences in our datasets. These are represented to our network with a combined vocab consisting of $50K$ word level tokens and $27$ parent tokens. The parent tokens come from the part of speech tags originally defined by the Penn Treebank \cite{PenTags} plus a root token. Following recent work in GEC \cite{JunczysDowmunt:2018, Chollampatt:2018tj}, the word level tokens are converted into sub-words using a Byte Pair Encoding (BPE) trained on the large Wikipedia dataset \cite{heinzerling2018bpemb}. The BPE segments rare words into multiple subwords, avoiding the post-processing of unknown words used in many existing GEC techniques.

We test our network on two GEC benchmarks: the commonly used NUCLE CoNLL 2014 task \cite{Ng:2013}, and the AESW dataset \cite{Daudaravicius:2016}. CoNLL 2014 training data contains $57K$ sentences extracted from essays written by non-native English learners. Following the majority of existing GEC work, we augment the small CoNLL dataset with the larger NIST Lang-8 corpus. The Lang-8 data contains $1.1M$ sentences of crowd sourced data taken from the Lang-8 website, making it noisy and of lower quality than the CoNLL data. We test on the $1,312$ sentences in CoNLL 2014 testing set and use the $1,381$ sentences of the CoNLL 2013 testing set as validation data. For evaluation we use the official CoNLL M2scorer algorithm to determine edits and compute precision and recall \cite{m2scorer:14}. 

We also explore the large AESW dataset. AESW was designed in order to train grammar error identification systems. However, it includes both incorrect and corrected versions of sentences, making it useful for GEC as well. AESW contains $1.2M$ training, $143K$ testing, and $147K$ validation sentences. The AESW data was taken from scientific papers authored by non-native speakers, and as such contains far more formal language than CoNLL.

\subsubsection{GEC Training}

For GEC we include a few additions to the training procedure described in Section \ref{sec:training}. First, we pre-train the network in two ways on $100M$ sentences from the large monolingual dataset provided by \cite{JunczysDowmunt:2016um}. We pre-train the decoder in our network as a language model, including all layers in the decoder except the encoder-decoder attention mechanism. We also pre-train the entire model as a denoising-autoencoder, using a source embedding word dropout of 0.4.

For loss we use the edit-weighted MLE objective defined by Junczys-Dowmunt et al. \cite{JunczysDowmunt:2018}:
\[ MLE(x,y) = - \sum_{t=1}^T \lambda(y_t) log P(y_t | x, y<t) \]
where $(x,y)$ are a training pair, and $\lambda(y_t)$ is $3$ if $y_t$ is part of an edit and 1 otherwise. We compute which tokens are part of an edit using the python Apted graph matching library \cite{AptedWeb, Apted1, Apted2}.

During beam-search we ensemble our networks with the monolingual language-model used for pre-training as per \cite{Xie:2016vi}:
\[ s(y|x) = \sum_{t=1}^T log P(y_t | x, y_{<t}) + \alpha log P_{lm}(y_t | y_{<t}) \] 
where $\alpha$ is chosen between $0$ and $1$ based on the validation set. Typically, we found an alpha of 0.15 performed best. Networks with $\alpha > 0$ are labeled with +Mon-Ens

\subsubsection{CoNLL 2014 Analysis}
 Results for CoNLL 2014 are provided in Table \ref{tab:conll}. Our Tree-Transformer achieves significantly higher recall than existing approaches, meaning we successfully repair more of the grammar errors. However, our precision is also lower which implies we make additional unnecessary edits. We attribute this drop to the fact that our method tends to generate examples which fit a structured grammar. Thus sentences with uncommon grammar tend to be converted to a more common way of saying things. An example of this effect is provided in table \ref{tab:conll_example}.

\subsubsection{AESW Analysis}
Results for AESW are provided in Table \ref{tab:aesw}. We achieve the highest to date F$0.5$ score on AESW, including beating out our own sequence-based Transformer model. We attribute this to the fact that AESW is composed of samples taken from submitted papers. The more formal language used in this context may be a better fit for the structured grammar used by our model.

\begin{table}
  \vspace{-2.0mm}
  \caption{CoNLL 2014 results}
  \label{tab:conll}
  \centering
  \begin{tabular}{ l l l l }
    \toprule
    Architecture & Precision & Recall & F$0.5$ \\
    \midrule
    \multicolumn{4}{c}{\textit{Prior State-of-the-art Approaches}} \\
    \midrule
    Chollampatt and Ng 2017. & 62.74 & 32.96 & 53.14 \\
    Junczys-Dowmunt and Grundkiewicz. 2016 & 61.27 & 27,98 & 49.49 \\
    \midrule
    \multicolumn{4}{c}{\textit{Prior Neural Approaches}} \\
    \midrule
    Ji et al. 2017 & - & - & 45.15 \\
    Schmaltz et al. 2017 & - & - & 41.37 \\
    Xie et al. 2016 & 49.24 & 23.77 & 40.56 \\
    Yuan and Briscoe. 2016 & - & - & 39.90 \\
    Chollampatt and Ng. 2018 & \textbf{65.49} & 33.14 & 54.79 \\
    Junczys-Dowmunt et al. 2018 & 63.0 & 38.9 & \textbf{56.1} \\
    \midrule
    \multicolumn{4}{c}{\textit{This Work}} \\
    \midrule
    Tree-Transformer & 57.39 & 28.12 & 47.50 \\
    Tree-Transformer +Mon & 58.45 & 30.42 & 49.35\\
    Tree-Transformer +Mon +Mon-Ens & 57.84 & 33.26 & 50.39\\
    Tree-Transformer +Auto & 65.22 & 30.38 & 53.05\\
    Tree-Transformer +Auto +Mon-Ens & 59.14 & \textbf{43.23} & 55.09\\
    \bottomrule
  \end{tabular}
\end{table}

 \begin{table}
  \tiny
  \vspace{-1.0mm}
  \caption{CoNLL 2014 Example Output}
  \label{tab:conll_example}
  \centering
  \begin{tabular}{ l | l }
    \toprule
    Input & In conclusion , we could tell the benefits of telling genetic risk to the carriers relatives overweights the costs . \\
    \midrule
    Labels & In conclusion , we \textbf{can see that} the benefits of telling genetic risk to the carrier\textbf{'s} relatives \textbf{outweighs} the costs . \\
    & In conclusion , we \textbf{can see that} the benefits of \textbf{disclosing} genetic risk to the carriers relatives \textbf{outweigh} the costs. \\
    & In conclusion , we \textbf{can see that} the benefits of \textbf{revealing} genetic risk to the carrier\textbf{'s} relatives \textbf{outweigh} the costs . \\
    \midrule
    Network & In conclusion , \textbf{it can be argued} that the benefits of \textbf{revealing} genetic risk \textbf{to one 's} relatives \textbf{outweighs} the costs . \\
    \bottomrule
  \end{tabular}
  \vspace{-2.5mm}
\end{table}

\begin{table}[h]
  \vspace{-1.0mm}
  \caption{AESW results}
  \label{tab:aesw}
  \centering
  \begin{tabular}{ l l l l l }
    \toprule
    Architecture & Precision & Recall & F$0.5$ \\
    \midrule
    \multicolumn{4}{c}{\textit{Prior Approaches}} \\
    \midrule
    Schmaltz et al. 2017 (Phrased-based) & - & - & 38.31 \\
    Schmaltz et al. 2017 (Word LSTM) & - & - & 42.78 \\
    Schmaltz et al 2017 (Char LSTM) & - & - & 46.72 \\
    \midrule
    \multicolumn{4}{c}{\textit{This Work}} \\
    \midrule
    Transformer (Seq to Seq) & 52.3 & 36.2 & 48.03 \\
    Tree-Transformer & 55.4 & 37.1 & \textbf{50.43} \\
    \bottomrule
  \end{tabular}
\end{table}

\section{Conclusion}
In this paper we introduced the Tree-Transformer architecture for tree-to-tree correction tasks. We applied our method to correction datasets for both code and natural language and showed an increase in performance over existing sequence-based methods. We believe our model achieves its success by taking advantage of the strong grammatical structure inherent in tree-structured representations. For the future we hope to apply our approach to other tree-to-tree tasks, such as natural language translation. Additionally, we intend to extend our approach into a more general graph-to-graph method.   
\newpage
\bibliography{tree_transformer}
\bibliographystyle{unsrt.bst}

\newpage
\begin{appendices}
\section{hyperparameters} \label{app:A}
Hyperparameters utilized are listed in tables \ref{tab:hyp_mod} and \ref{tab:hyp_tra}. Default hyperparameters are listed at the top of each table. A blank means the run utilized the default hyperparameter.  Explanation of hyperparemeters follows.

\begin{itemize}
  \item \textbf{N} - number of layers
  \item \textbf{$d_{model}$} - size of sublayer outputs - see \cite{Vaswani:2017ul}
  \item \textbf{$d_{ff}$} - size of inner layer in TDB/FF for Tree-Transformer/Transfomer
  \item \textbf{$h$} - Number of attention heads
  \item \textbf{$d_k$} - Size of keys in attention mechanism
  \item \textbf{$d_v$} - Size of values in attention mechanism
  \item \textbf{$p_{drop}$} - Dropout probability between sub-layers
  \item \textbf{$p_{dattn}$} - Dropout probability on attention mechanism - see \cite{Vaswani:2017ul}
  \item \textbf{$p_{dff}$} - Dropout probability on inner layer of TDB/FF for Tree-Transformer/Transfomer
  \item \textbf{$p_{des}$} - Source embedding word dropout probability
  \item \textbf{$p_{det}$} - Target embedding word dropout probability
  \item \textbf{$\epsilon_{ls}$} - Label Smoothing $\epsilon$
  \item \textbf{lr} - Learning Rate, We use isr learning rate as per \cite{Vaswani:2017ul}. As such this learning rate will never be fully met, maximum learning rate depends upon warmup.
  \item \textbf{warmup} - number of steps for linearly LR warmup
  \item \textbf{train steps} - total number of steps for training
  \item \textbf{Mon} - Initialized from monolingual pre-trained network
  \item \textbf{Auto} - Initialized from autoencoder pre-trained network
  \item \textbf{Mon-Ens} - Ensemble trained network with monolingual netword during beam search as per \cite{Xie:2016vi} 
  \item \textbf{EW-MLE} - Use edit-weight MLE objective function as per \cite{JunczysDowmunt:2018}
  \item \textbf{Time (Hours)} - Total training time
 \end{itemize}
 
 \begin{table}
  \small
  \caption{Model Parameters}
  \label{tab:hyp_mod}
  \centering
  \begin{tabular}{ l | l l l l l l l l l l l l }
    \toprule
    Architecture & N & $d_{model}$ & $d_{ff}$ & $h$ & $d_k$ & $d_v$ & $p_{drop}$ & $p_{dattn}$ & $p_{dff}$ & $p_{des}$ & $p_{det}$ & $\epsilon_{ls}$  \\
    \midrule
    default & 6 & 512 & 2048 & 8 & 64 & 64 & 0.3 & 0.1 & 0.3 & 0.2 & 0.1 & 0.1 \\
    \midrule
    \multicolumn{13}{c}{\textit{Sate IV}} \\
    \midrule
    LSTM & 4 & 1024 & N/a & \\
    Transformer & \\
    Tree-Transformer & \\
    \midrule
    \multicolumn{13}{c}{\textit{GEC Pretraining}} \\
    \midrule
    Monolingual & \\
    Autoencoder & & & & & & & & & & 0.4 & & \\
    \midrule
    \multicolumn{13}{c}{\textit{Conll 2014}} \\
    \midrule
    Tree-Transformer & \\
    Tree-Transformer +Mon & \\
    Tree-Transformer +Mon +Mon-Ens& \\
    Tree-Transformer +Auto& \\
    Tree-Transformer +Auto +Mon-Ens & \\
    \midrule
    \multicolumn{13}{c}{\textit{Aesw}} \\
    \midrule
    Transformer & & & & & & & & & & \\
    Tree-Transformer & \\
    \bottomrule
    
  \end{tabular}
\end{table}

\begin{table}
  \small
  \caption{Training Parameters}
  \label{tab:hyp_tra}
  \centering
  \begin{tabular}{ l | l l l | l l l l | l}
    \toprule
    Architecture & lr & warmup & train steps & Mon & Auto & Mon-Ens & EW-MLE & Time (Hours)  \\
    \midrule
    default & $d_{model}^{-0.5}$ & 4000 & 100k & - & - & - & - &\\
    \midrule
    \multicolumn{9}{c}{\textit{Sate IV}} \\
    \midrule
    LSTM & & & & & & & & 40 \\
    Transformer & & & & & & & & 18 \\
    Tree-Transformer & & & & & & & & 22 \\
    \midrule
    \multicolumn{9}{c}{\textit{GEC Pretraining}} \\
    \midrule
    Monolingual & & & 500k & & & & & 38 \\
    Autoencoder & & & 500k & & & & & 50 \\
    \midrule
    \multicolumn{9}{c}{\textit{Conll 2014}} \\
    \midrule
    Tree-Transformer & & 16000 & & & & & 3 & 26 \\
    Tree-Transformer +Mon & & 16000 & & \checkmark & & & 3 & 26 \\
    Tree-Transformer +Mon +Mon-Ens & & 16000 & & \checkmark & & \checkmark & 3 & 26 \\
    Tree-Transformer +Auto & & 16000 & & & \checkmark & & 3 & 26 \\
    Tree-Transformer +Auto +Mon-Ens & & 16000 & & & \checkmark & \checkmark & 3 & 26 \\
    \midrule
    \multicolumn{9}{c}{\textit{Aesw}} \\
    \midrule
    Transformer & & 16000 & & \checkmark & & \checkmark & 3 & 19 \\
    Tree-Transformer & & 16000 & &  \checkmark & & \checkmark & 3 & 25 \\
    \bottomrule
  \end{tabular}
\end{table}

\end{appendices}

\end{document}